\documentclass[APA,LATO1COL]{WileyNJD-v2}
\usepackage{moreverb}
\usepackage{amsmath}
\usepackage{amsfonts}
\usepackage{amssymb}
\usepackage{amsthm}
\usepackage{enumitem}
\usepackage{fancyhdr}
\usepackage{float}
\usepackage{bbm}
\usepackage{ulem}
\usepackage{setspace}
\usepackage[export]{adjustbox}
\usepackage{lineno}
\usepackage{tikz}
\usepackage{booktabs}
\usepackage{multirow}
\usepackage{subcaption}
\usepackage{natbib}

\usepackage{graphicx}
\usepackage{hyperref}%
\usepackage{hypernat}

\hypersetup{breaklinks=true,
            pdfauthor={},
            pdfkeywords = {},
            pdftitle={},
            colorlinks=true,
            citecolor=blue,
            urlcolor=blue,
            linkcolor=magenta,
            pdfborder={0 0 0}}

\newcommand\BibTeX{{\rmfamily B\kern-.05em \textsc{i\kern-.025em b}\kern-.08em
T\kern-.1667em\lower.7ex\hbox{E}\kern-.125emX}}

\articletype{Article}%

\received{<day> <Month>, <year>}
\revised{<day> <Month>, <year>}
\accepted{<day> <Month>, <year>}

\newcommand{\R}{\mathbbm{R}}
\newcommand{\g}{\gamma}

\usepackage{xcolor}

\newcommand{\change}[1]{\color{black}{#1}\normalcolor}

\begin{document}

\title{Dimensionality Reduction using Elastic Measures}

\author[1]{J. Derek Tucker*}
\author[2]{Matthew T. Martinez}
\author[3]{Jose M. Laborde}

\authormark{Tucker \textsc{et al}}

\address[1]{\orgdiv{Statistical Sciences}, \orgname{Sandia National Laboratories}, \orgaddress{\state{NM}, \country{United States}}}

\address[2]{\orgdiv{}, \orgname{Boeing}, \orgaddress{\state{NM}, \country{United States}}}

\address[3]{\orgdiv{}, \orgname{Moffitt Cancer Center}, \orgaddress{\state{FL}, \country{United States}}}

\corres{*J. Derek Tucker, \email{jdtuck@sandia.gov}}

\presentaddress{P.O. Box 5800 MS 0829 \\
Albuquerque, NM 87185}

\abstract[Abstract]{With the recent surge in big data analytics for hyper-dimensional data there is a renewed interest in dimensionality reduction techniques. In order for these methods to improve performance gains and understanding of the underlying data, a proper metric needs to be identified. This step is often overlooked and metrics are typically chosen without consideration of the underlying geometry of the data. In this paper, we present a method for incorporating elastic metrics into the t-distributed Stochastic Neighbor Embedding (t-SNE) and Uniform Manifold Approximation and Projection (UMAP). We apply our method to functional data, which is uniquely characterized by rotations, parameterization, and scale. If these properties are ignored, they can lead to incorrect analysis and poor classification performance. Through our method we demonstrate improved performance on shape identification tasks for three benchmark data sets (MPEG-7, Car data set, and Plane data set of Thankoor), where we achieve 0.77, 0.95, and 1.00 F1 score, respectively.}


\jnlcitation{\cname{%
\author{J.D. Tucker}, 
\author{M.T. Martinez}, and 
\author{J.M. Laborde}} (\cyear{2022}), 
\ctitle{Dimensionality Reduction using Elastic Measures}, \cjournal{Stat}, \cvol{2022.aa.bb}.}

\maketitle

\section{Introduction} \label{introduction}
With the ever increasing number of observations and expanding number of variables in modern datasets, the use of dimensionality reduction as a first step for training machine learning models has gained a renewed interest. Methods such as the t-distributed Stochastic Neighbor Embedding (t-SNE)~\citep{tSNE} and the Uniform Manifold Approximation and Projection (UMAP)~\citep{McInnes:2018d50}, have enabled improved classification performance over methods such as principal component analysis (PCA) and factor analysis. This is due to their ability to accurately embed complicated data structures into small dimensional spaces by computing the distances between observations in higher dimensional space. The choice of distance function is typically left to the user, where the default metric is the Euclidean distance, which is deficient in higher dimensional space. It has been shown that the UMAP is sensitive to the choice of distance metric and can effect the results significantly~\citep{McInnes:2018d50}. 

When computing the distances between observations, a metric which is symmetric, isometric, and obeys the triangle inequality is preferred. 
These properties lead to pairwise distance computations that are independent of input order and can be degenerate for some data types~\citep{srivastava-etal-JASA:2011,tucker2013}.
For example, the Euclidean distance is not a proper distance on functional data in $\mathbb{R}^1$ under phase variation~\citep{tucker2013}, the space of open and closed curves~\citep{srivastava2016functional}, and any trajectory on a Riemmanian manifold \citep{su2014statistical}.

In this paper, we demonstrate that selecting a distance metric that is consistent with the data topology drastically improves the performance of both t-SNE and UMAP on functional data. 
We apply this method to a clustering of curves tasks in $\mathbb{R}^n$, where we choose our distance metric such that it measures the \textit{shape} of these curves by  accounting for rotation, scale, and parameterization. We utilize the elastic shape framework \citep{srivastava2016functional}, which yields two metrics that properly measure distance in both shape and amplitude, and provides a distance from one shape to another in sampling variability. When applied to the space of open curves, a Riemmanian manifold, we are able to obtain an initialization point for dimensionality reduction that is consistent with the properties of the data. We also apply this framework to any trajectory on a Riemmanian manifold using the square-root velocity transformation.
Additionally, we can change out the cost-function in t-SNE from Kullabck-Liebler divergence to a proper distance on the space of PDFs which provides performance gains depending on the balance of precision and recall desired. 

This paper is organized as follows. In Section~\ref{sec:esa} we provide an overview of the Elastic Shape Analysis (ESA) framework.
In Section~\ref{sec:dimension} we review both t-SNE and UMAP and provide modification in using a proper distance for functional data and a modification to the cost-function to t-SNE.
Section~\ref{sec:results} provides results on two data sets, one being a set of shapes from the MPEG7, shapes of cars extracted from video, and fighter jet shapes, and finally, we provide a discussion and concluding remarks in Section~\ref{sec:discussion}.

\section{Elastic Shape Analysis}\label{sec:esa}
Elastic Shape Analysis (ESA) is a collection of techniques for registering and analyzing functional data, using the process of phase-amplitude separation.
From the separated components, statistical analysis can be performed on the individuals phase and amplitude components \citep{srivastava2011, kurtek2011Signal, tucker2013}. 
In this section we provide a review of ESA and refer the reader to \citep{srivastava2011, kurtek2011Signal, tucker2013, srivastava2016functional} for a complete overview.

Phase and amplitude represent two orthogonal components of a function's shape, which are properties of a function that remain invariant to shape-preserving transformations (rotation, translation, scaling, and phase)~\citep{srivastava2016functional,dyrden:2016}. 
The phase component represents the \textit{horizontal} variability within trajectories or reparameterization, where as the amplitude component represents the \textit{vertical} variability independent of phase.
In order to analyze these components the need to be separated through alignment, \citet{srivastava2011} demonstrated that the Fisher-Rao metric is a proper distance and provides this separation.

To compute the alignment using the Fisher-Rao metric, the Square Root Velocity Function (SRVF) \citep{srivastava2011} is used for registration of functions in $\mathbbm{R}^n$. 
For real-valued curves, the SRVF performs a bijective mapping of the real-valued curve $f$ to its normalized gradient $f' / \sqrt{|f'|}$. 
The registration of two real-valued curves is performed by elastically deforming the domain of one function such that the $L^2$ distance between the SRVFs of the two curves is minimized.
The amount of elastic deformation required is measured by the phase distance (Section \ref{sec:phase}), while the residual distance between the SRVF, post registration (rotation, and scaling), defines the amplitude distance between them (Section \ref{sec:rn}). 
Together they are known as elastic distances. 
By registering SRVFs instead of trajectories directly, the phase and amplitude distances become proper metrics and are invariant to shape-preserving transformations.

\subsection{Amplitude distance for $\R^n$ valued functions} \label{sec:rn}
Let $F_{\mathbbm{R}^n} = \{f:[0, 1] \mapsto \mathbb{R}^n, f \text{ differentiable} \}$ be the class of differentiable trajectories on $[0, 1]$ mapping to $\R^n$ with $n \geq 2$\footnote{We are focusing here on $n\geq 2$ and similar metrics are available for $n=1$.}. In higher dimensional Euclidean space ($n \geq 2)$, the scale, rotation, and phase of a trajectory need to be accounted for in order to isolate its shape. Scale variability is removed by standardizing each trajectory to have unit length, i.e., each trajectory is divided by the $L^2$ norm of its gradient:
\[
f(t) \mapsto f(t) / ||f'|| \quad \forall t \in [0, 1],
\]
where $||f'|| = \sqrt{\int_0^1 f'(t)^2dt}$. 

Rotation variability is accounted for using the space of rotation matrices, the Special Orthogonal Group $SO(n)$ and is defined as the group of orthogonal matrices with a determinant of one. For trajectories in $\mathbbm{R}^2$, $SO(2)$ defines the rotations around a point and in $\mathbbm{R}^3$ $SO(3)$ defines rotations around a line. The action of $SO(n)$ on a trajectory $f$ is denoted as $O(f)$ and is defined pointwise as 
\[ O(f) = \{O f(t) : \forall t \in [0, 1] \},\] where $O f(t)$ represents a standard matrix multiplication of the $n \times n$ matrix $O$ and the $n \times 1$ vector $f(t)$. See \cite{popov1994orthogonal} for more details on the Orthogonal groups and their properties. 
The optimal rotation matrix in $SO(n)$ is found alongside the optimal reparameterization when computing the amplitude distance.

Phase variability in $\mathbbm{R}^n$ is represented using the space $\Gamma=\{\gamma: I \rightarrow I|\lbrace\gamma, \gamma^{-1}\rbrace \in C^1(I), \dot \gamma >0\}$ of all positive slope diffeomorphisms of the unit interval. 
Together with the operation of composition, $\Gamma$ is a group and is defined with respect to the domain $I = [0, 1]$ . 
\change{The properties of the diffeomorphisms is what allows the bending and stretching described above and why we use the term \textit{elastic}.}
The reparameterization of a curve $f$ by a warping function $\gamma\in\Gamma$ is denoted as the operation $f\circ\gamma$.

We desire a distance that is invariant to simultaneous reparameterizations, which allows us to use the SRVF as the following transformation on trajectories $f \in F_R^n$:
The SRVF is defined as the following transformation on trajectories $f \in F_R^n$:
\begin{equation} \label{def:srvf}
q_f(t) = \frac{f'(t)}{\sqrt{|f'(t)|}},
\end{equation}
where $|f'(t)|$ is the absolute value of $f'$ at $t$.
This transformation maps trajectories onto the $L^2$ ball of radius one and is a bijective mapping, up to an additive constant \citep{srivastava2016functional}. 
The reparmeterization of $q_f(t)$ is then $(q_f(t),\gamma) = (q_f(t)\circ\gamma)\sqrt{\dot{\gamma}}$.

Since the space of SRVFs is the $L^2$ ball then the norm on the SRVFs is the arc length distance between points on the sphere:
\begin{equation} \label{eqn:Rnmetric}
    d(q_f, q_g) = \arccos{{\int_0^1 \langle q_f(t), q_g(t) \rangle dt}},
\end{equation}
where $q_f = SRVF(f)$ and $q_g = SRVF(g)$ for two trajectories $f, g \in F_{R^n}$ and $\langle q_f(t), q_g(t) \rangle$ denotes the inner product of the vectors $q_f(t), q_g(t)$. 
To convert this into an amplitude distance we need to place $f$ and $g$ in phase and rotation with each other. 
This optimization is summarized as the following amplitude distance on $\mathbbm{R}^n$ valued trajectories \citep{srivastava2011}:
\begin{equation} \label{eqn:Rndist}
(f, g) = \inf_{\g \in \Gamma, O \in SO(n)} \arccos{{\int_0^1 \langle q_f(t), q_{O(g \circ \gamma)}(t)\rangle dt}},
\end{equation}
where $q_f$ and $q_{O(g \circ \g)}$ denote the SRVF's of $f$ and $O(g \circ \g)$ respectively.

\subsection{Phase Space} \label{sec:phase}
We will define the phase space using diffeomorphic mapping. 
The phase space of the unit interval $[0, 1]$ as defined above is 
\[
\Gamma = \{\g : [0, 1] \mapsto [0, 1] \ | \ \g(0) = 0 \text{, } \g(1) = 1 \text{, } \g \text{ is a diffeomorphism}\}.
\]
This diffeomorphic constraint gives rise to the notion of elasticity because the elements of $\Gamma$, i.e., phase functions, can only smoothly stretch and contract portions of the unit interval so that it maps back to itself. 
Phase is generally thought of as the representation of a trajectory because any trajectory with domain $[0, 1]$ can be \textit{warped} by a phase function to appear differently. 
The amplitude will be taken to be those features of a trajectory that remain unchanged under any possible warping. 

\subsubsection{Phase distance} \label{phase}
To define phase distance we use the optimal $\gamma$ that defines the amplitude distance.
The phase space $\Gamma$ is a nonlinear manifold with no known geometry so we use the SRVF to map $\Gamma$ to a known geometry \citep{tucker2013}. 
Phase functions are positive for all $t \in [0, 1]$ and $||q_{\gamma}|| = \int_0^1 \sqrt{\gamma'(t)}^2 dt = 1$, so the SRVF maps $\Gamma$ onto the positive orthant of a unit Hilbert Sphere. 
Thus the phase distance is defined as 
\begin{equation} \label{phsdistance}
        d_{p}(\gamma_1, \gamma_2) = \arccos{{\int_0^1 \langle \psi_1, \psi_2 \rangle dt}},
\end{equation}
where $\psi=\sqrt{\dot{\gamma}}$ and $\langle \cdot, \cdot \rangle$ is the inner product between the two vectors \citep{srivastava2016functional}. 
The metric $d_{p}(\gamma_1, \gamma_2)$ is essentially measuring the amount of elastic deformation needed to compare the shapes of $f_1$ and $f_2$, when $\gamma_1=id$ is the identity warping.

\section{Dimension Reduction}\label{sec:dimension}
In dimensionality reduction, we are interested in finding a low dimensional embedding space, $Y$, for the high dimensional space, $X$, such that $Y$ is as similar to $X$ as possible and that the representation of the reduced space facilitates better data analysis (e.g., classification with good class separation). 
For the dimensionatliy reduction to work we want to define the similarities between two objects $x_i$ and $x_j$ in $X$ and their mapping to $Y$. Recently, two methods have become highly utilized in the literature for performing dimensionality reduction; t-SNE and UMAP. 
In the following sections we will give a brief overview of the two methods. 

\subsection{t-SNE: t-distributed Stochastic Neighbor Embedding}\label{sec:t-SNE}
A t-distributed Stochastic Neighbor Embedding (t-SNE)~\citep{tSNE} embedding defines probabilities proportional to the similarity of the objects $x_i$ and $x_j$, and are defined as
\[p_{j|i} = \frac{\exp(-d(x_i,x_j)^2/2\sigma_i^2)}{\sum_{k\neq i} \exp(-d(x_i,x_k)^2/2\sigma_i^2)},\]
were $p_{i|i}=0$, the $\sum_j p_{j|i} = 1 \forall i$, $d(x_i,x_j)$ is the Euclidean distance between the two objects (as defined in~\citep{tSNE}). The similarity between the two objects is the conditional probability that $x_i$ is similar to $x_j$ in proportion to the probability density under a Gaussian distribution centered at $x_i$.

The total probability or similarity is defined as
\[p_{ij} = \frac{p_{j|i}+p_{i|j}}{2N}\]
where t-SNE aims to learn a $d$-dimensional mapping, $Y$, that reflects the similarities $p_{ij}$. The similarities between $y_i$ and $y_j$ are defined as:
\[q_{ij} = \frac{(1+d(y_i,y_j)^2)^{-1}}{\sum_k\sum_{l\neq k}(1+d(y_i,y_j)^2)^{-1}},\]
for which the student t-distribution measures these similarities in the mapping space. 

The locations of the points ($y_i$) in the lower dimensional space is determined by minimizing the Kullback-Leibler divergence of the distribution $P$ from the distribution $Q$, i.e., 
\[KL(P||Q) = \sum_{i\neq j} p_{ij} \log\left(\frac{p_{ij}}{q_{ij}}\right).\]
This minimization is performed using gradient descent and is the mapping between $X$ and $Y$.

\subsection{UMAP: Uniform Manifold Approximation and Projection}\label{sec:umap}
Uniform Manifold Approximation and Projection (UMAP) approximates a manifold by constructing a fuzzy simplicial set representation, which is performed on high dimensional data ($X$) and on a low dimensional representation ($Y\in\mathbb{R}^{d}$). The selected representation is the one that optimizes the cross entropy between the high and low dimensional spaces.

UMAP is similar to the approach of t-SNE, which constructs a probability distribution over pairs of high-dimensional objects. In t-SNE similar objects have a high probability of being picked, whereas dissimilar points have an extremely small probability of being picked.  t-SNE defines a similar probability distribution over the points in the low-dimensional map, and minimizes the Kullback-Leibler divergence between the two distributions.

In UMAP, the high dimensional similarities, $v_{j|i}$, are local fuzzy simplicial set memberships based on smooth nearest-neighbor (NN) distances $v_{j|i}$ from $x_{i}\in X$ to one of its $k$ distinct nearest-neighbors $x_{j}\in X$ 
\[v_{j|i} = \exp[(-d(x_{i},x_{j})+\rho_{i})/\sigma_{i}]\] where $d(x_{i},x_{j})$ is the Euclidean distance on the learned manifold. 
The parameter $\rho_{i}$ is the distance to nearest neighbor and $\sigma_{i}$ is a normalizing constant. 
The symmetrization are used to produce an undirected graph structure representing the 1-dimensional frame of the fuzzy simplicial set, is
carried out by fuzzy set union using the probabilistic t-conorm \[v_{ij}=(v_{j|i}+v_{i|j})-v_{j|i}v_{i|j}.\] 
The graph defined by the $v_{ij}$ is then embedded into a low dimensional space $Y$, where the dimension is prescribed as a parameter. 
The low dimensional similarities between the projections $y_{i}$ and $y_{j}$, of $x_{i}$ and $x_{j}$, into $Y$ via the initial embedding are given by \[w_{ij}=\left(1+a||y_{i}-y_{j}||_{2}^{2b}\right)^{-1}\] 
where $a$ and $b$ are user defined and a gradient descent procedure is used to find them. 
The defaults for UMAP are $a\approx1.929$ and $b\approx0.7915$ \citep{McInnes:2018d50}.

The cost function that is optimized to find the embedding is
\[C_{UMAP}=\sum_{i\neq j}v_{ij}\log\left(\frac{v_{ij}}{w_{ij}}\right)+(1-v_{ij})\log\left(\frac{1-v_{ij}}{1-w_{ij}}\right)\]
which penalizes discrepancies in the relative distributions of similarities in $X$ and $Y$.

\subsection{Elastic Modifications}
To perform dimensionality reduction, it is critical that the distance, $d(x_i,x_j)$, is appropriate for the underlying data. 
In this work, we perform dimensionality reduction on curves that lie in $\mathbb{R}^n$.
We then can utilize the distance defined in Equation~\ref{eqn:Rndist} for measuring the distance between two open curves, where the distance is symmetric, isometric, and obeys the triangle inequality, unlike the Euclidean distance. 
Additionally, we perform clustering on the phase-variability of the data using the phase distance in Equation~\ref{phsdistance}.

The cost function for t-SNE uses the Kullback-Leibler divergence which is not a metric in the space of probability density functions (PDFs), however, different divergences have been studied for t-SNE~\citep{ftsne, bunte:2012}.
Depending on the the choice of divergence a different balance is sought between precision and recall of the embedding.
For example, in \cite{ftsne}, the authors showed that the Hellinger distance, an extrinsic metric on the transformed space of PDFs, balances precision and recall while penalizing small embeddings and preserving neighborhood sizes. 
For a complete review of the effect of various diverges the reader is referred to \citep{ftsne}.
The use of an intrinsic metric instead of an extrinsic metric should be done as it is not an approximation.
The Fisher-Rao metric is an intrinsic on the space of PDFs and can be used in favor of the Hellinger or Kullback-Leibler.

We give a brief review of the geometry of the space of PDFs but refer the reader to \cite{kurtek:2015} for a more in-depth review. 
We focus here on univariate densities on $[0,1]$, but the theory can easily be extend for all finite dimensional densities.
Let $\mathcal{P}=\{p:[0,1]\rightarrow\mathbb{R}_{\geq 0} | \int_0^1 p(x)\,dx = 1\}$ as the space of all PDFs and we define the tangent space $T_p(\mathcal{P}) = \{d_p:[0,1]\rightarrow\mathbb{R} | \int_0^1 d_p(x)p(x)\,dx=0\}$. 
The inner-product on this space, known as the Fisher-Rao Riemmanian metric \citep{rao:45,Kass97} is given by:
\[\langle\langle d_{p_1},d_{p_2}\rangle\rangle_p = \int_0^1 d_{p_1}(x)d_{p_2}(x)\frac{1}{p(x)}\,dx\]

\cite{Cencov82} showed that this metric is invariant to re-parameterizations. 
However, the metric is difficult to compute, requiring numerical methods and large computations. 
\cite{bhattacharya-43} proposed a convenient square-root transformation with simplifies the computation of the metric and is the transformed space where the Hellinger distance is an extrinsic metric. 
Define the transformation $\phi(p) = \sqrt{p}$ as the square-root density of PDF $p$, which is similar to SRVF described previously. 
The corresponding space of SRDs is $\Psi = \{\sqrt{p}:[0,1]\rightarrow\mathbb{R}_{\geq 0} | \int_0^1p(x)\,dx = 1\}$ which the positive orthant of the Hilbert Sphere.
Since the geometry is known in this transformed space we can define the geodesic distance between two PDFs $p_1,p_2 \in \mathcal{P}$ under the Fisher-Rao metric using the SRDS $\sqrt{p_1},\sqrt{p_2}\in\Psi$ is the great circle defined as
\begin{equation} \label{eq:frpdf}
	d_{FR}(p_1,p_2) = \cos^{-1}\left(\int_0^1\sqrt{p_1(x)}\sqrt{p_2(x)}\,dx\right).
\end{equation}

Therefore, we can replace the Kullback-Leibler divergence in t-SNE with a proper metric on the space of PDFs using Equation~\ref{eq:frpdf} similar to divergences done in \cite{ftsne, bunte:2012}.
The use of an intrinsic metric is more natural than an extrinsic method as it is the exact distance between two PDFs and is invariant to re-parametrizations of the PDFs
We will denote this modification to t-SNE as \change{\textit{Fisher-Rao t-SNE} or frt-SNE and will compare t-SNE, UMAP, and Fisher-Rao t-SNE} using the elastic metrics on the space of curves in $\mathbb{R}^n$.

\section{Results}\label{sec:results}
In this section we provide results on two data set. 
The first data set will be analyzing a set of shapes where we have 65 shape classes from the MPEG-7 Core Experiment CE-Shape-1 Test Set~\citep{mpeg7}. 
This data set is used for benchmarking shape matching algorithms and includes binary images grouped into categories by their content, not their appearance or shape.
The second is the fighter plane shapes and the car shapes from \cite{thankoor:2007} and includes part of the MPEG-7 data set.

\subsection{MPEG7}\label{sec:mpeg7}
\begin{figure}[htb]
  \centering
  \input{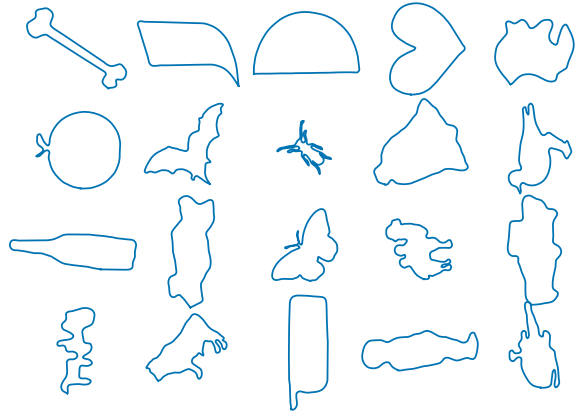}\vspace{-.3in}
  \caption{Example shapes from the MPEG-7 shape database.}
  \label{fig:mpeg7}
\end{figure}

In this section, we classify real world curve data taken from the MPEG-7 database~\citep{mpeg7}. 
The full database has 1300 shape samples, 20 shapes for each class, and 65 shape classes. 
An example of some of the shapes from the MPEG-7 bases is presented in Figure~\ref{fig:mpeg7}. 
Note the differences in rotations and scales of each of the shapes which complicates the analysis.

\begin{figure}[htb]
  \centering
  \begin{subfigure}[t]{.5\textwidth}
    \centering
    \includegraphics{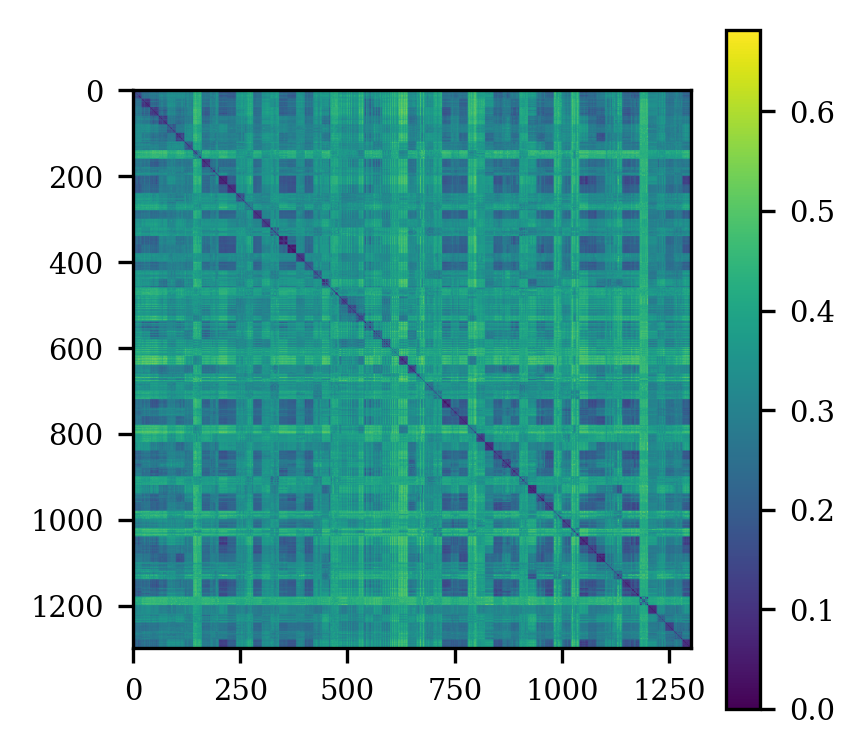}
    \caption{Elastic Distance}\label{fig:dist_fr}
  \end{subfigure}%%
  \begin{subfigure}[t]{.5\textwidth}
    \centering
    \includegraphics{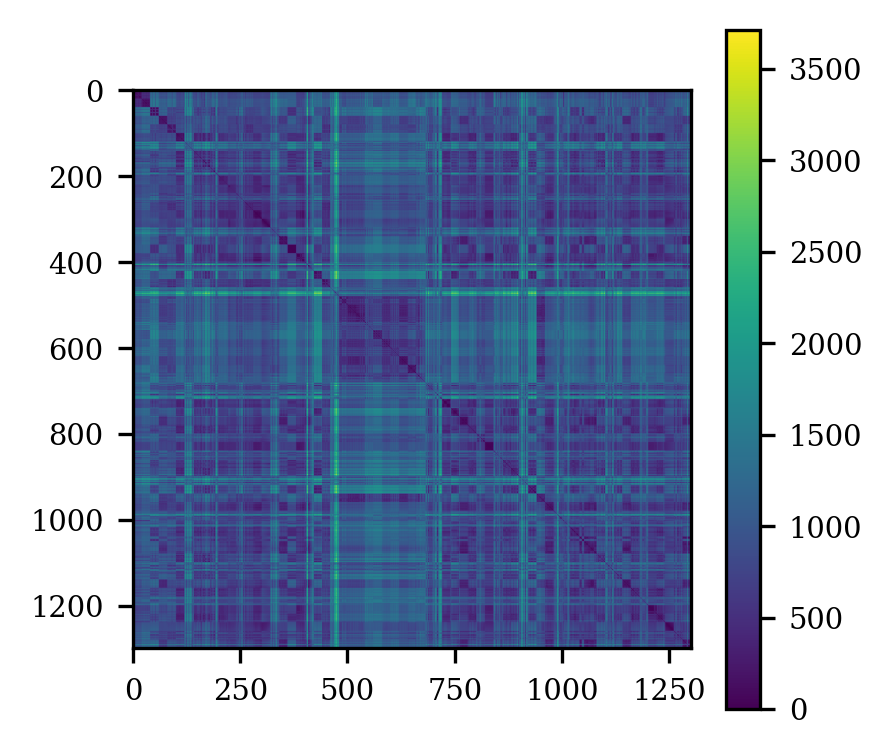}
    \caption{Euclidean Distance}\label{fig:dist_euc}
  \end{subfigure}
  \caption{MPEG7 Distance Matrices.}
  \label{fig:mpeg7_dist}
\end{figure}

Using the shape distance in Definition~\ref{eqn:Rndist} we can compute the pairwise distance matrix between all shapes in the data set. 
Figure~\ref{fig:dist_fr} provides the distance matrix using the elastic distance and for comparison we computed the distance metric using the Euclidean distance and is provided in Figure~\ref{fig:dist_euc}. 
\change{The euclidean distance computed does not remove rotation or scale variation as this is the default distance in the original implementation of t-SNE and UMAP. Additionally, even if we remove this variation as part of the Euclidean distance it is still not a proper distance on the space of open curves. We will utilize this Euclidean distance for all results in this section.}
Through inspection of the distance matrices there is a definite difference in structure and visually there is a noticeable large amount of clustering in the elastic distance where the Euclidean distance is only showing roughly three large groups.

\begin{figure}[htbp]
  \centering
  \begin{subfigure}[t]{.5\textwidth}
    \centering
    \includegraphics{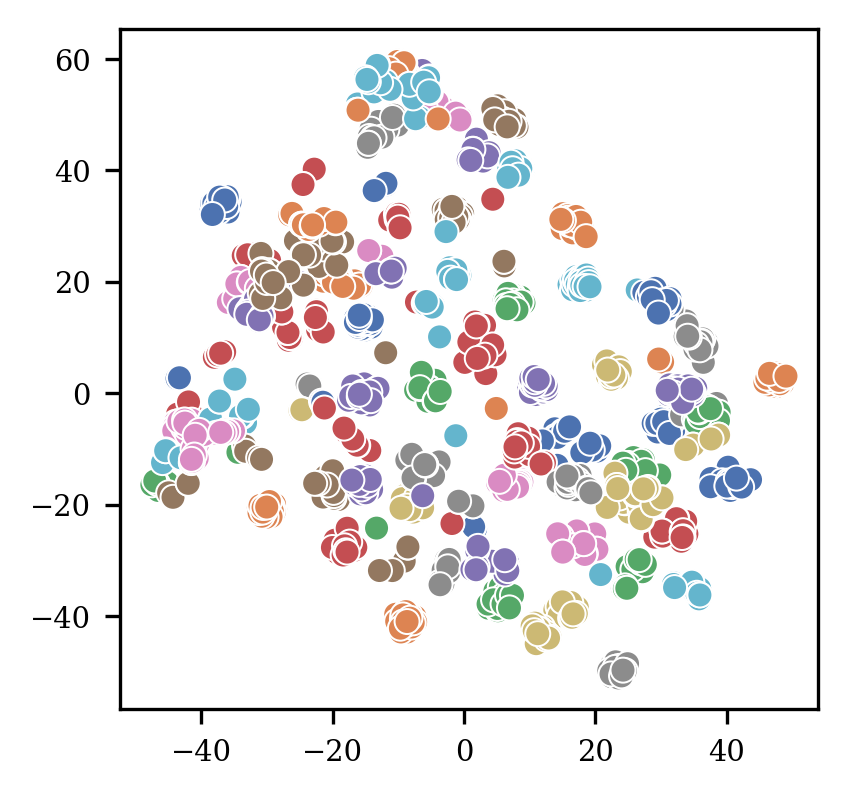}
    \caption{Elastic Distance}\label{fig:mpeg7_t-SNE_fr}
  \end{subfigure}%%
  \begin{subfigure}[t]{.5\textwidth}
    \centering
    \includegraphics{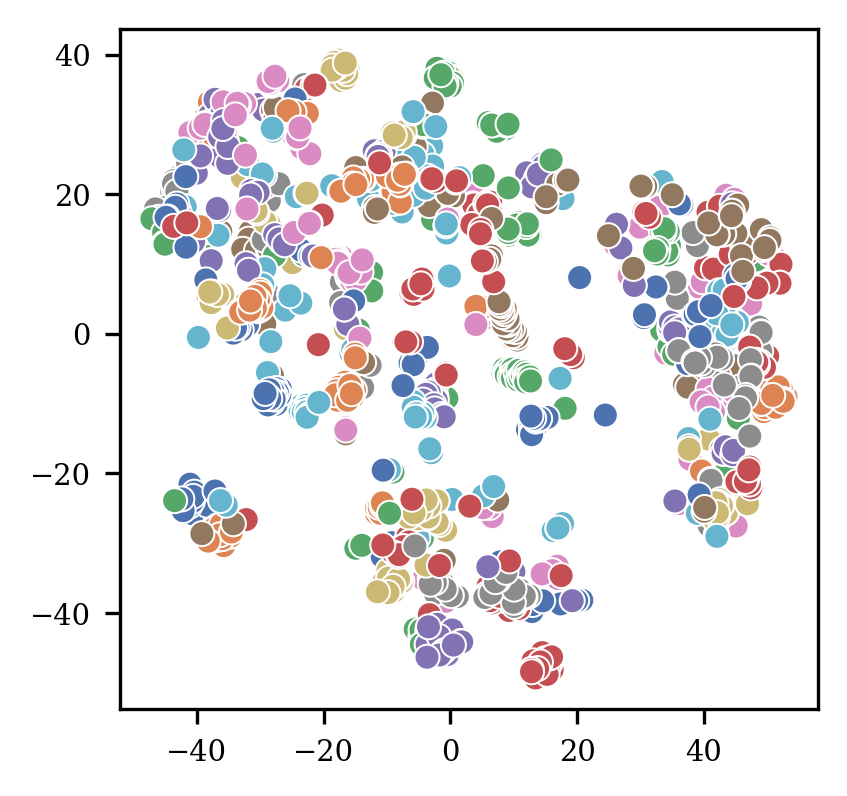}
    \caption{Euclidean Distance}\label{fig:mpeg7_t-SNE_euc}
  \end{subfigure}
  \caption{MPEG7 t-SNE Projection into two dimensions.}
  \label{fig:mpeg7_t-SNE}
\end{figure}

\begin{figure}[htbp]
  \centering
  \begin{subfigure}[t]{.5\textwidth}
    \centering
    \includegraphics{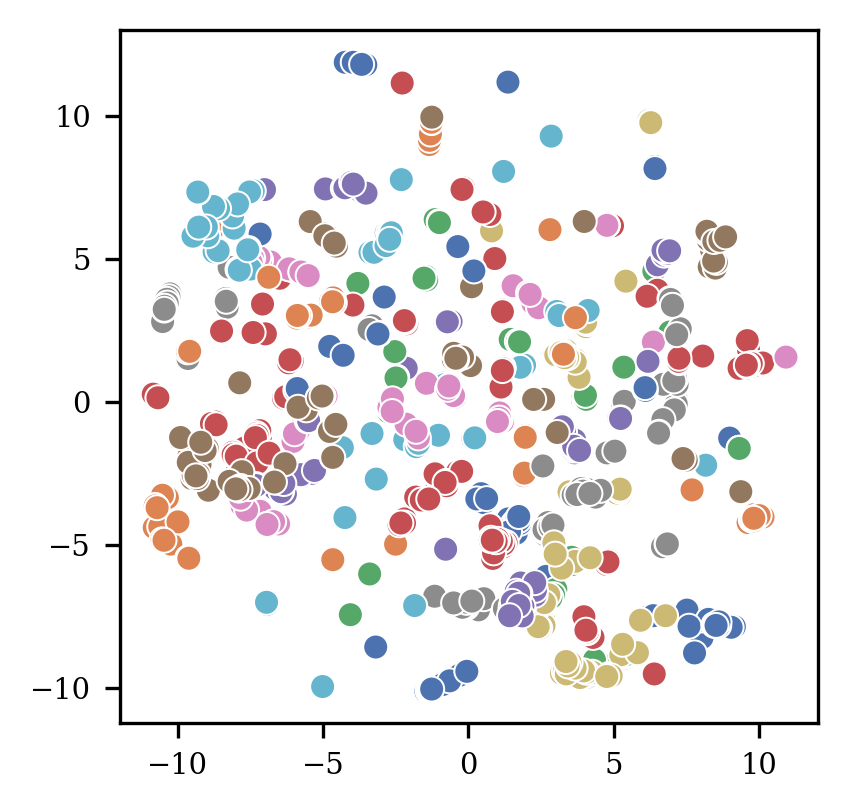}
    \caption{Elastic Distance}\label{fig:mpeg7_et-SNE_fr}
  \end{subfigure}%%
  \begin{subfigure}[t]{.5\textwidth}
    \centering
    \includegraphics{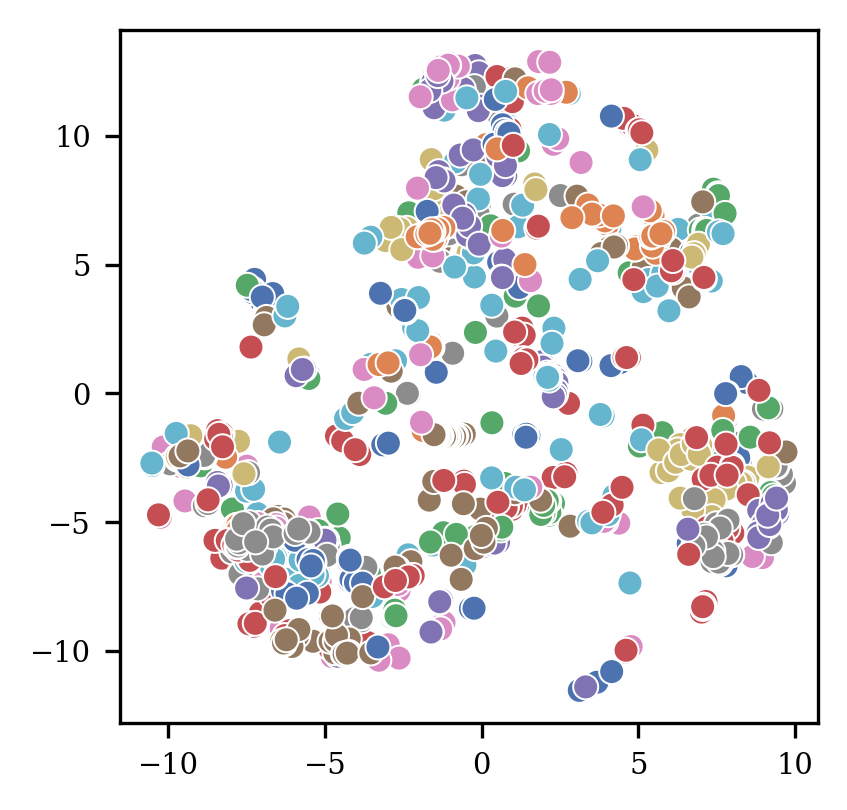}
    \caption{Euclidean Distance}\label{fig:mpeg7_et-SNE_euc}
  \end{subfigure}
  \caption{MPEG7 Fisher-Rao t-SNE Projection into two dimensions.}
  \label{fig:mpeg7_et-SNE}
\end{figure}

\begin{figure}[htbp]
  \centering
  \begin{subfigure}[t]{.5\textwidth}
    \centering
    \includegraphics{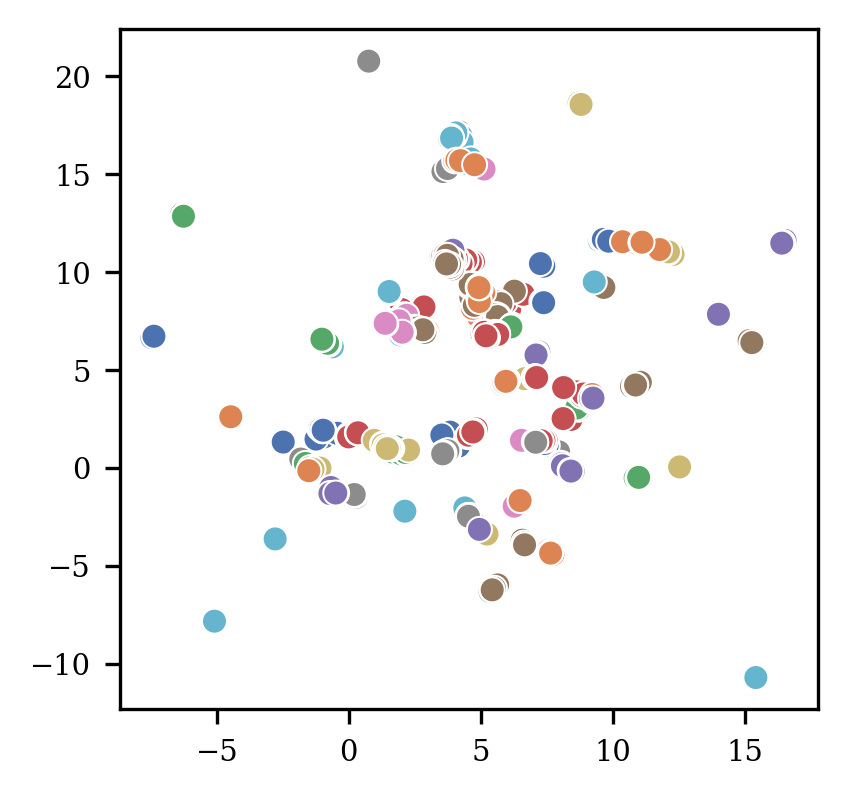}
    \caption{Elastic Distance}\label{fig:mpeg7_umap_fr}
  \end{subfigure}%%
  \begin{subfigure}[t]{.5\textwidth}
    \centering
    \includegraphics{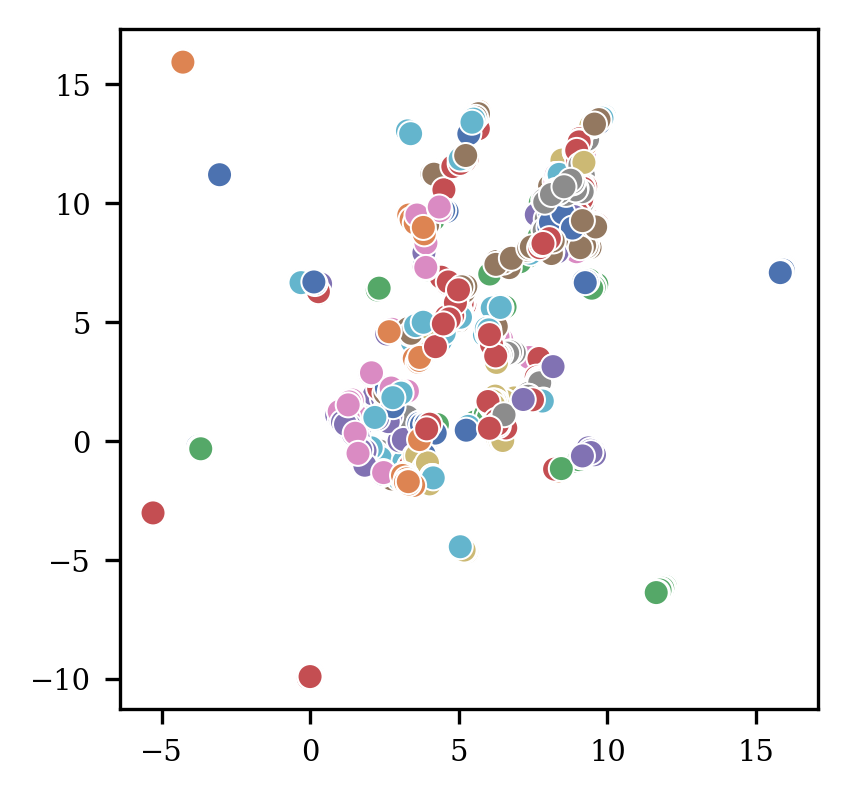}
    \caption{Euclidean Distance}\label{fig:mpeg7_umap_euc}
  \end{subfigure}
  \caption{MPEG7 UMAP Projection into two dimensions.}
  \label{fig:mpeg7_umap}
\end{figure}

We computed both t-SNE, frt-SNE, and UMAP projecting the data onto two components using the elastic distance and the Euclidean distance for comparison.
Figure~\ref{fig:mpeg7_t-SNE} provides the t-SNE results with panel (a) using the elastic distance and panel (b) using the Euclidean distance.
There is a distinct clustering of each of the classes and separability of the classes versus the Euclidean distance. 
Figure~\ref{fig:mpeg7_et-SNE} provides the Fisher-Rao t-SNE results with panel (a) using the elastic distance and panel (b) using the Euclidean distance.
Comparing to standard t-SNE using the Kullback-Leibler divergence we see a slightly better separation of classes and as mentioned in above we see neighborhood sizes are more preserved.
Figure~\ref{fig:mpeg7_umap} provides the UMAP results with panel (a) using the elastic distance and panel (b) using the Euclidean distance.
Again, we see a distinct clustering of the classes and better separability using a proper metric for the space of open curves. 

\begin{figure}[htb]
  \centering
  \includegraphics{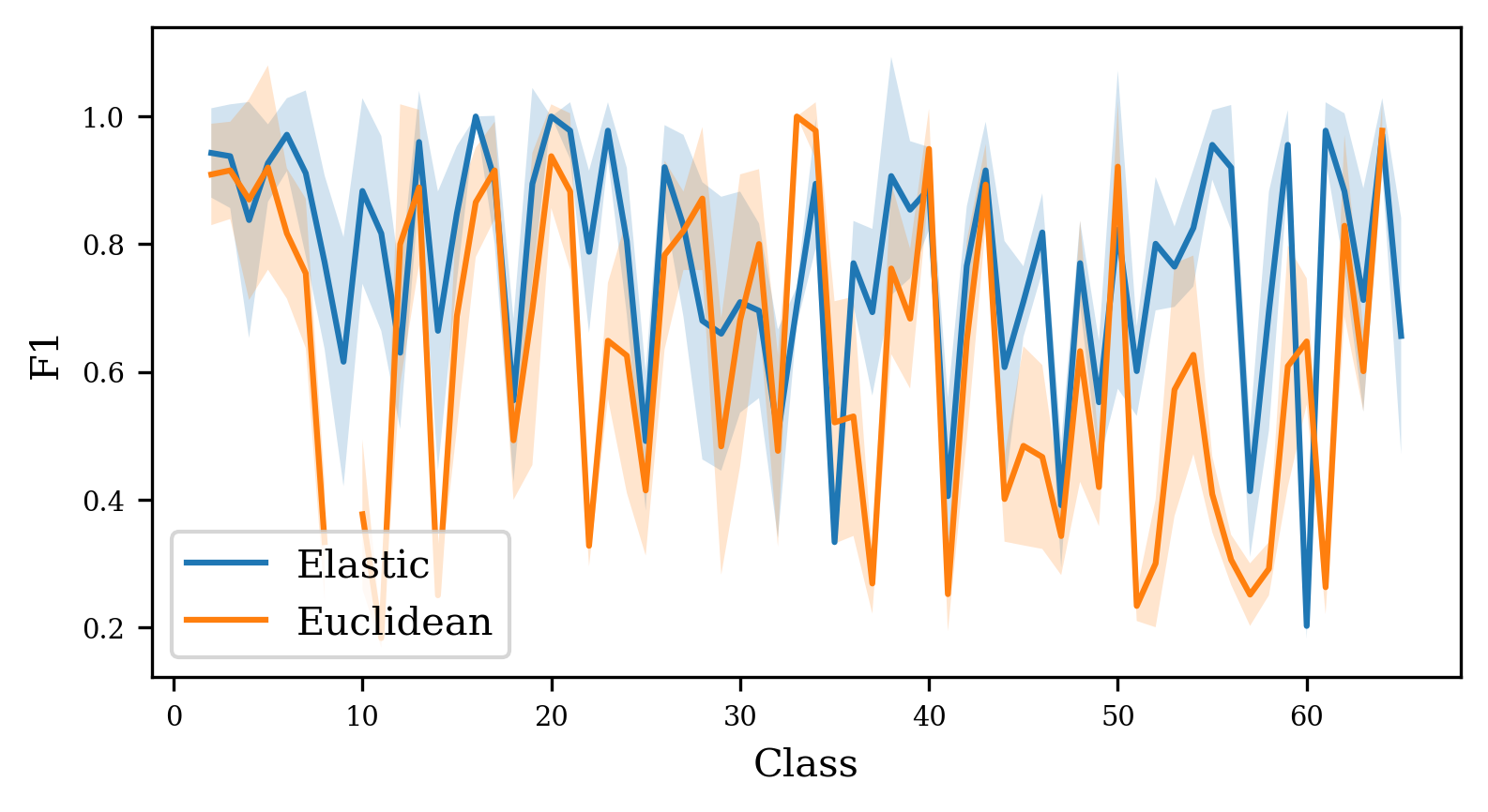}
  \caption{Average F1 Score MPEG7}
  \label{fig:mpeg7_f1}
\end{figure}

\begin{figure}[htb]
  \centering
  \begin{subfigure}[t]{.5\textwidth}
    \centering
    \includegraphics{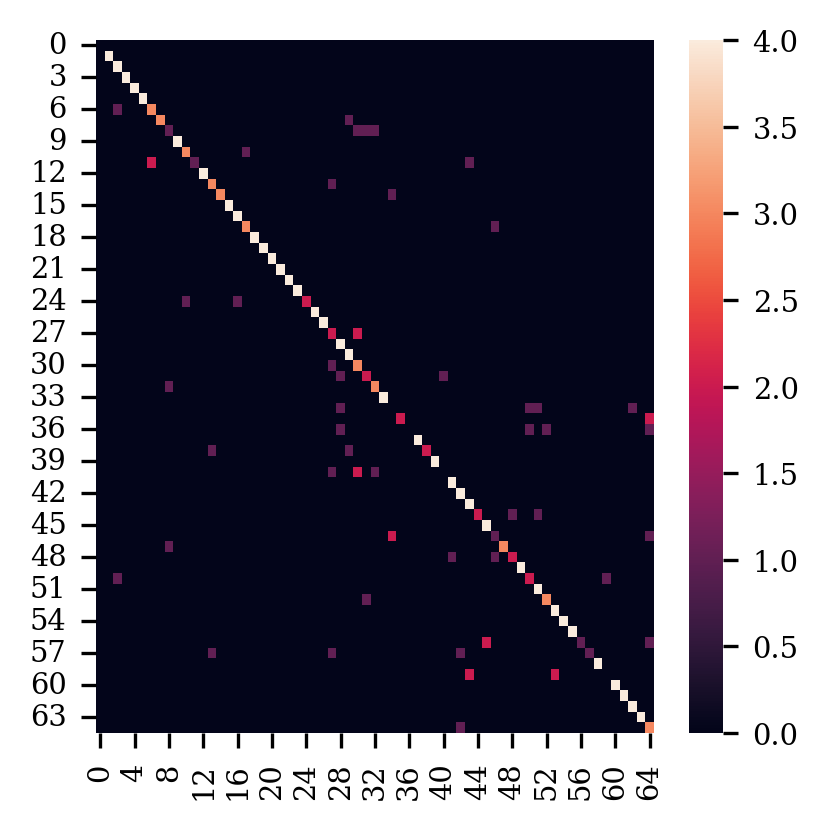}
    \caption{Elastic Distance}\label{fig:mpeg7_conf_fr}
  \end{subfigure}%%
  \begin{subfigure}[t]{.5\textwidth}
    \centering
    \includegraphics{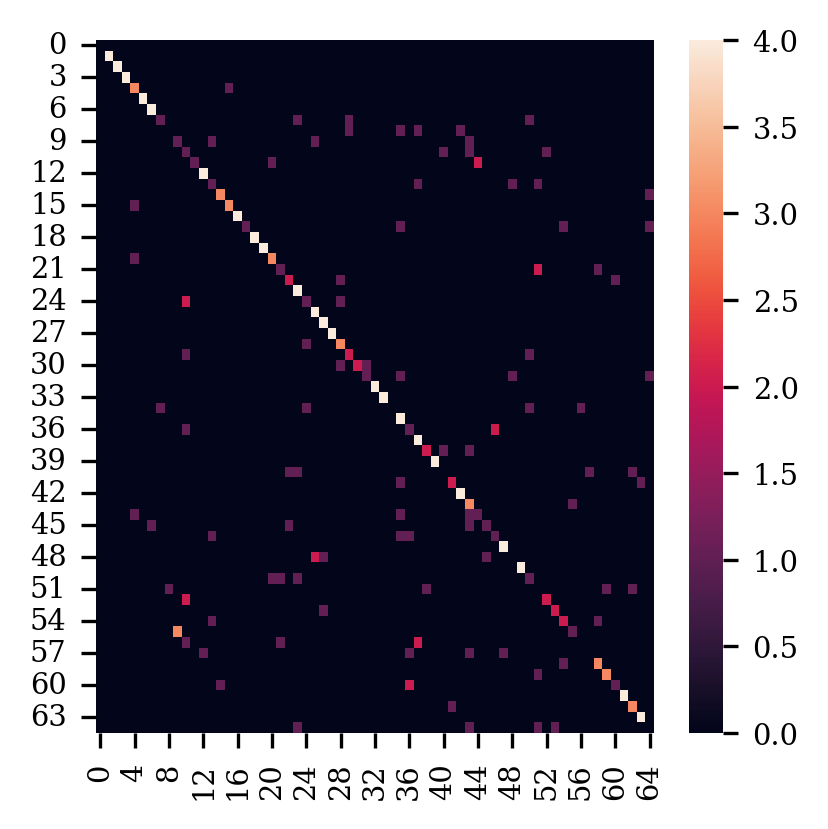}
    \caption{Euclidean Distance}\label{fig:mpeg7_conf_euc}
  \end{subfigure}
  \caption{MPEG7 confidence matrices.}
  \label{fig:mpeg7_conf}
\end{figure}

To add a more complete picture to the embedding we trained a Random Forest classifier \citep{breiman} using 5-fold cross-validation. 
Since, t-SNE was not originally defined for out-of-sample data we will focus on UMAP.
In the following section, for the full data analysis we will utilize parametric t-SNE \citep{vandermatten:2009,policar:2021} which allows for out-of-sample data. 
Figure~\ref{fig:mpeg7_f1} provides the average F1 score across folds with the standard deviation shaded, using the elastic metric and the Euclidean metric. 
Overall, for each class the F1 score is larger for each class and maintains on average over 0.8 while on average of the Euclidean is 0.6. 
Figure~\ref{fig:mpeg7_conf} provides the confidence matrix for the elastic distance in panel (a) and the Euclidean metric in panel (b).
As was noted from the F1 score, the elastic distance for most of the classes does extremely well with few classes being misspecified. 
The Euclidean distance does have trouble with a few of the classes with more than a few miss classifications. 

\subsection{Shape Database}\label{sec:shape_results}
In this section, we look at the overall classification performance utilizing UMAP and parametric t-SNE as a dimensionality reduction tool and then performing classification using a Random Forest classifier.
Again we will compare the results against using the standard Euclidean metric in both dimension reduction techniques.
For this study we will utilize the fighter jet shapes, car shapes, and the full MPEG7 database described and analyzed in the previous section.

The fighter airplane shape database includes Mirage, Eurofighter, F-14, Harrier, F-22, and F-15 shapes. 
Since the F-14 has two possible shapes, one when its wings are closed and another when its wings are opened, the total number of shape classes is seven, where there are 30 shapes for each class for a total of 210 examples.
The database was created from digital pictures of die-cast replica of the airplanes and were segmented to extract the contour.

The car shapes were generated using the method described in \cite{thankoor:2005} where contours of vehicles were extracted from video clips. 
There are four classes in the group: sedan, pickup, minivan, and SUV and each have 30 samples for each class. 
The data set shows larger within-class variation, as shapes of vehicles of different makes and models vary and some contours are distorted due to the shadow also being extracted. 
This data set contains more real world variability in applying shape classification metrics. 

\begin{figure}[htb]
  \centering
  \begin{subfigure}[t]{.7\textwidth}
    \centering
    \includegraphics[width=\textwidth]{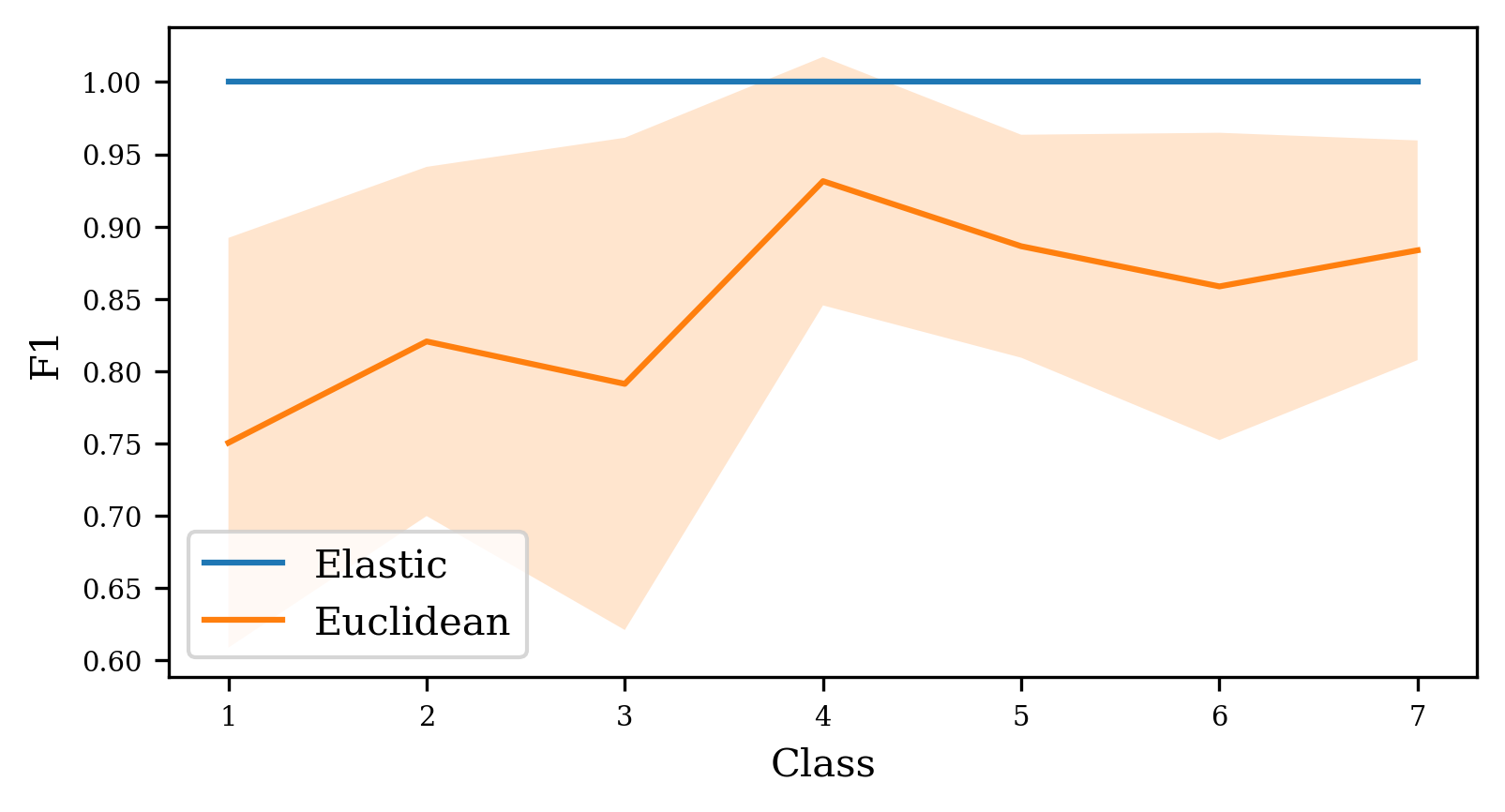}
    \caption{Planes}\label{fig:f1_planes}
  \end{subfigure}\\
  \begin{subfigure}[t]{.7\textwidth}
    \centering
    \includegraphics[width=\textwidth]{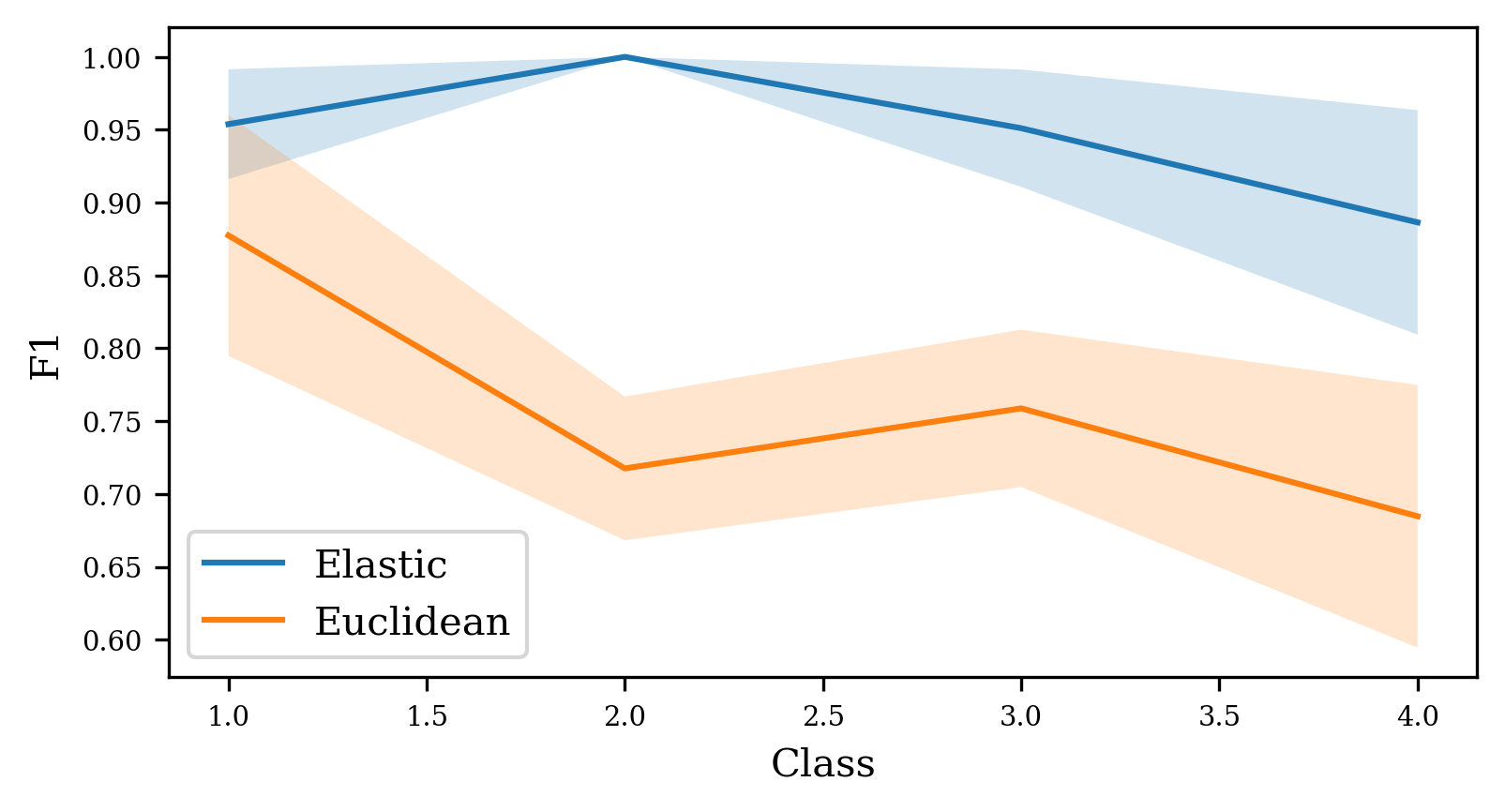}
    \caption{Cars}\label{fig:f1_cars}
  \end{subfigure}
  \caption{F1 score for each class for the plane and car data sets using UMAP and Random Forest.}
  \label{fig:mshape_conf}
\end{figure}

Figure~\ref{fig:mshape_conf} presents the F1 score for each class for the plane data set in panel a) and the car data set in panel b).
A Random Forest classifier was trained from the UMAP embedding using the designated distance.
The shaded region shows the variability across the 5-fold cross-validation.
In both data sets the elastic metric vastly outperforms the the Euclidean metric for all of the classes and shows less variability across the folds.
The increase in performance is directly related to utilizing a proper distance and a metric that is invariant to scale, rotation, and re-parameterization of the shape.
Being invariant to these actions is critical in any shape analysis/recognition task.

Table~\ref{tabl:class_mcc} presents a look at the class average F1 score and class averaged Matthew's Correlation Coefficient (MCC) for all three data sets: planes, cars, and MPEG7.
The average was taken across classes and across folds from the 5-fold cross-validation of a Random Forest Classifier with the designated metric and dimensionality reduction technique.
Panel a) presents the metrics utilizing UMAP, panel b) presents the metrics utilizing t-SNE, and panel c) presents the metrics using Fisher-Rao t-SNE.
For UMAP the elastic method outperforms the Euclidean metric for both the F-score and MCC and for the easier planes data set was perfect. 
For t-SNE we achieve similar results where the elastic metrics out perform the Euclidean metric.
The same story over improvement in using the metrics when for Fisher-Rao t-SNE. 
We see an improvement across all data sets when using the Fisher-Rao metric as the cost-function for t-SNE.
The performance overall is degraded using t-SNE over UMAP as UMAP computes a more separable lower dimensional representation. 
Additionally, as has been well known across the literature, t-SNE is more computationally expensive as the number of examples in the data set increases. 
The one benefit of t-SNE over UMAP is the statistical interpretation which can be used for uncertainty quantification.

\begin{table}[htbp]
  \centering
  \begin{subtable}[t]{.5\textwidth}\centering
    \begin{tabular}{lllll}
        \toprule
        \multirow{2}{*}{data set} & \multicolumn{2}{c}{Elastic} & \multicolumn{2}{c}{Euclidean}\\
        \cmidrule{2-3} \cmidrule{4-5} \\
        {} & F-score & MCC & F-score  & MCC \\
        \midrule
        Planes & 1.000  & 1.000 & 0.8461  & 0.8286 \\
        Cars & 0.9478 & 0.9367 & 0.7596 & 0.6897 \\
        MPEG-7 & 0.7694 & 0.7587 & 0.6210 & 0.5807 \\
        \bottomrule
    \end{tabular}
    \caption{UMAP}
  \end{subtable}\\%%
  \vspace{.3in}
  \begin{subtable}[t]{.5\textwidth}\centering
    \begin{tabular}{lllll}
        \toprule
        \multirow{2}{*}{data set} & \multicolumn{2}{c}{Elastic} & \multicolumn{2}{c}{Euclidean}\\
        \cmidrule{2-3} \cmidrule{4-5} \\
        {} & F-score & MCC & F-score  & MCC \\
        \midrule
        Planes & 0.9554  & 0.9400 & 0.7826  & 0.7524 \\
        Cars & 0.8130 & 0.7662 & 0.7370 & 0.6655 \\
        MPEG-7 & 0.5833 & 0.4614 & 0.5686 & 0.4213 \\
        \bottomrule
    \end{tabular}
    \caption{t-SNE}
  \end{subtable}\\%%
  \vspace{.3in}
  \begin{subtable}[t]{.5\textwidth}\centering
    \begin{tabular}{lllll}
        \toprule
        \multirow{2}{*}{data set} & \multicolumn{2}{c}{Elastic} & \multicolumn{2}{c}{Euclidean}\\
        \cmidrule{2-3} \cmidrule{4-5} \\
        {} & F-score & MCC & F-score  & MCC \\
        \midrule
        Planes & 0.9904  & 0.9891 & 0.8038  & 0.7811 \\
        Cars & 0.8240 & 0.7962 & 0.7461 & 0.6775 \\
        MPEG-7 & 0.5990 & 0.5282 & 0.5281 & 0.4252 \\
        \bottomrule
    \end{tabular}
    \caption{Fisher-Rao t-SNE}
  \end{subtable}
  \caption{Class Averaged Metrics for Shape data set.}
  \label{tabl:class_mcc}
\end{table}

Lastly, to examine performance across multiple classifiers we compared the performance of the Random Forest (RF) against the Gradient Boosted Trees (GBT) \citep{friedman} and K-nearest neighbors classifiers (KNN) \citep{thomas}.
The purpose of this comparison is to demonstrate that the method is independent of the classifier and to show the power is in choosing the proper metric for the geometry of the data under consideration.
Table~\ref{tabl:mult_class_mcc} presents the class averaged F1 score and MCC for the three classifiers for the MPEG-7 data set.
For all three classifiers under both UMAP and t-SNE we see an improvement in classification performance when using the elastic metric over the Euclidean. 
Across the classifiers, UMAP has the most consistent performance, while t-SNE obtains the highest under KNN, the performance under the other classifiers is quite reduced.
The increase in performance between the two metrics is attributed to picking a metric that is a \textit{proper} distance for the data being classified. 

\begin{table}[htbp]
  \centering
  \begin{subtable}[t]{.5\textwidth}\centering
    \begin{tabular}{lllll}
        \toprule
        \multirow{2}{*}{data set} & \multicolumn{2}{c}{Elastic} & \multicolumn{2}{c}{Euclidean}\\
        \cmidrule{2-3} \cmidrule{4-5} \\
        {} & F-score & MCC & F-score  & MCC \\
        \midrule
        RF & 0.7694 & 0.7587 & 0.6210 & 0.5807 \\
        GBT & 0.6926 & 0.6209 & 0.6190 & 0.4716 \\
        KNN & 0.7942 & 0.7813 & 0.6987 & 0.6387 \\
        \bottomrule
    \end{tabular}
    \caption{UMAP}
  \end{subtable}\\%%
  \vspace{.3in}
  \begin{subtable}[t]{.5\textwidth}\centering
    \begin{tabular}{lllll}
        \toprule
        \multirow{2}{*}{data set} & \multicolumn{2}{c}{Elastic} & \multicolumn{2}{c}{Euclidean}\\
        \cmidrule{2-3} \cmidrule{4-5} \\
        {} & F-score & MCC & F-score  & MCC \\
        \midrule
        RF & 0.5990 & 0.5282 & 0.5281 & 0.4252 \\
        GBT & 0.5409 & 0.3432 & 0.4814 & 0.2525 \\
        KNN & 0.8480 & 0.8352 & 0.7227 & 0.6660 \\
        \bottomrule
    \end{tabular}
    \caption{t-SNE}
  \end{subtable}\\%%
  \vspace{.3in}
  \begin{subtable}[t]{.5\textwidth}\centering
    \begin{tabular}{lllll}
        \toprule
        \multirow{2}{*}{data set} & \multicolumn{2}{c}{Elastic} & \multicolumn{2}{c}{Euclidean}\\
        \cmidrule{2-3} \cmidrule{4-5} \\
        {} & F-score & MCC & F-score  & MCC \\
        \midrule
        RF & 0.5833 & 0.4614 & 0.5686 & 0.4213 \\
        GBT & 0.4402 & 0.3133 & 0.4610 & 0.3270 \\
        KNN & 0.6212 & 0.5706 & 0.5538 & 0.4685 \\
        \bottomrule
    \end{tabular}
    \caption{Fisher-Rao t-SNE}
  \end{subtable}
  \caption{Class Averaged Metrics for MPEG-7 across multiple classifiers}
  \label{tabl:mult_class_mcc}
\end{table}

\section{Discussion}\label{sec:discussion}
We have proposed a new flexible elastic metric for functional data utilizing state-of-the art dimensionality reduction methods. 
We have demonstrated its advantages over the current methods using current shape analysis benchmarking data set. 
Unlike the current cross-sectional metrics the elastic method accurately estimates the mean of the underlying data generating mechanism and handles rotation and parametrization as a nuisance accordingly.
It is imperative to choose a metric that is respective of the underlying variability of the data and is respective of the geometry.
Doing so will give advantages in multiple machine learning tasks.
We have presented the metric for open and closed shapes in $\mathbb{R}^n$, but equivalent metrics exists for functions in $\mathbb{R}^1$ \citep{tucker2013}, surfaces~\citep{kurtek:2017}, and images~\citep{xie:2016}. 

\section*{Acknowledgments}
This paper describes objective technical results and analysis. Any subjective views or opinions that might be expressed in the paper do not necessarily represent the views of the U.S. Department of Energy or the United States Government.
This work was supported by the Laboratory Directed Research and Development program at
Sandia National Laboratories, a multi-mission laboratory managed and
operated by National Technology and Engineering Solutions of Sandia,
LLC, a wholly owned subsidiary of Honeywell International, Inc., for the
U.S. Department of Energy's National Nuclear Security Administration
under contract DE-NA0003525.

\bibliography{reference}

\end{document}